\documentclass{article}
\usepackage{arxiv}
\usepackage[utf8]{inputenc} 
\usepackage[T1]{fontenc}    
\usepackage{hyperref}       
\usepackage{url}            
\usepackage{booktabs}       
\usepackage{amsfonts}       
\usepackage{nicefrac}       
\usepackage{microtype}      
\usepackage{lipsum}
\usepackage{subcaption}     
\usepackage{graphicx}
\graphicspath{{./images/}}
\usepackage{textcomp}
\usepackage{xcolor}
\usepackage{multirow}       
\usepackage{float}
\usepackage{yfonts}
\usepackage{amsmath}
\usepackage{amssymb}
\usepackage{amsthm}         
\usepackage{algorithm}
\usepackage{algpseudocode}
\usepackage{enumerate}
\usepackage{bm}
\usepackage{comment}
\usepackage{enumitem}
\usepackage{soul}
\usepackage{tabularx}
\usepackage[most]{tcolorbox}
\usepackage[authoryear]{natbib}
\newtheorem{mydefinition}{Definition}
\newtheorem{myremark}{Remark}
\newtheorem{myexample}{Example}
\floatname{algorithm}{Procedure}
\def\BibTeX{%
  {\rm B\kern-.05em{\sc i\kern-.025em b}\kern-.08em%
  T\kern-.1667em\lower.7ex\hbox{E}\kern-.125emX}}

\title{Agent-Based Detection and Resolution of Incompleteness and Ambiguity in
Interactions with Large Language Models}

\author{
Riya Naik \\
BITS Pilani, K K Birla Goa Campus \\
\texttt{p20210056@goa.bits-pilani.ac.in}
\And
Ashwin Srinivasan \\
BITS Pilani, K K Birla Goa Campus \\
\texttt{ashwin@goa.bits-pilani.ac.in}
\AND
Swati Agarwal \\
PandaByte Innovations Pvt Ltd \\
\texttt{agrswati@ieee.org}
\And
Estrid He \\
RMIT University \\
\texttt{estrid.he@rmit.edu.au}
}
  
\begin{document}

\maketitle
\begin{abstract}

Many of us now treat Large Language Models (LLMs) as
modern-day oracles,  asking it almost any kind of question. However, consulting an LLM does
not have to be a single-turn activity.
But long multi-turn interactions can get tedious, if it
is simply to clarify contextual information that can be arrived at through reasoning.
In this paper, we examine the use of agent-based architecture
to bolster LLM-based Question-Answering (QA) systems with additional reasoning capabilities.
We examine the automatic resolution of potential incompleteness or ambiguities in questions by transducers implemented using LLM-based agents. We focus on several benchmark datasets that are known
to contain questions with these deficiencies to varying degrees.
We equip two different LLMs (GPT-3.5-Turbo and Llama-4-Scout) with agents that
act as specialists in detecting and resolving--at least partially--deficiencies of incompleteness and ambiguity. The agents are implemented as zero-shot ReAct agents. Rather than producing an answer in a single step, the language model now decides between 3 actions: (a) classify; (b) resolve; and (c) answer. Action (a) decides if the question is incomplete, ambiguous, or normal; and action (b) determines if any deficiencies identified can be resolved. Action (c) answers the resolved form of the question. We compare the use of LLMs with and without the use of agents with these components.
Our results show these benefits of agents with transducer:
(1) A shortening of the length of interactions with human; 
(2) An improvement in the answer quality; and
(3) Explainable resolution of deficiencies in the question. On the negative side, we find while the approach may result in additional LLM invocations and, in some cases, increased latency. But on tested datasets, the benefits outweigh the costs except when questions already have sufficient context. Suggesting the agent-based approach could be a useful mechanism to harness the power of LLMs to develop more robust QA systems.
\end{abstract}

\section{Introduction}
Conversational AI systems have long been anticipated, both in fiction and real life.
While the focus in fiction has inevitably been on the
modality of delivery (voice), the content has not been quite as demanding
(``Tea. Earl Grey. Hot.''), perhaps reflecting the state-of-the-art of natural language
processing at the time of conception. However, this latter aspect has recently
been undergoing rapid re-evaluation with the advent of Large Language
Models (LLMs). LLM-based technology is increasingly part of the
design of systems involving some form of human interaction. These systems control how we access information, interact with support services, and use devices \cite{su-etal-2025-llm}.  It is therefore not unexpected that a
significant amount of academic and industrial
research and development is invested in a better understanding
of interactions with LLMs. 

Early applications of LLMs were confined to prompt-response 
interactions, functioning as static text generators without capabilities for planning 
or external interactions.  
Techniques such as chains, ReAct, and few-shot prompting introduced limited step-by-step processing, enabling LLMs to perform sequential reasoning \cite{stahl-etal-2024-exploring}.  
However, these methods often lack the flexibility required to handle complex, dynamic tasks \cite{kucharavy2024fundamental}.  
The recent emergence of goal-based autonomous agents has
enhanced the utility of LLMs. These frameworks integrate LLMs within systems
capable of multi-step reasoning, context maintenance across various functions, and 
interfacing with external environments like APIs and data sources 
\cite{fan-etal-2025-ai}. Consequently, LLMs can now be part of
software that can plan, observe, decide, and act, which makes
them a dynamic component in the design of intelligent systems \cite{sreedhar2025simulating}.

This paper is concerned with the design of intelligent question-answering (QA)
systems. It is now evident that modern LLMs can store very large amounts of context-information accumulated during the course of an interaction (input token-lengths are now millions of tokens, and only expected to increase \cite{achiam2023gpt}). Techniques such as in-context learning that provide
feedback in the form of clarificatory
instances and domain knowledge can greatly assist an LLM in retrieving relevant answers to queries \cite{dong-etal-2024-survey,chitale-etal-2024-empirical}. The amount of such clarificatory information can often be surprisingly small \cite{li-2025-review}. LLMs have the mechanisms
necessary for conducting interactive conversations \cite{daryanto2025conversate}. But, a question of some conceptual importance is this: when do LLMs need
interactive feedback, and how should this be provided? There are 3 ways to approach such a question. First, from a mathematical
point-of-view, by assessing if there
is an information-theoretic gap in what is known and what is asked. The LLM may
then need clarification to reduce this uncertainty (used in a mathematical sense). 
Secondly, is a technological approach. In this, we
could attempt to see if current LLM technology (software or hardware) make it impossible for the LLM to distinguish
between some aspects of the problem (for example, positional encoding makes it impossible to see the
question is about sets and not sequences). Finally, a behavioral
approach, in which we try to assess the conditions for feedback by examining the
question, and possibly the context that preceded it. It is this third category
that is of interest to us in this paper. 
We focus on whether augmenting an LLM with agents capable of
identifying and resolving problems with the question can improve the
QA performance of an LLM. Specifically, we focus on
agents for identifying 2 kinds of problems with a question: whether
it is incomplete; and whether it is ambiguous. We motivate the use
of agents through a simple conceptual model of QA
as a form of communication between agents. In this model, incompleteness
and ambiguity are defined in an ideal sense using the response of
an oracle. Practically, such an oracle is usually unavailable, and the
agents are intended to act as proxies with limited background information
(when compared to the oracle).

\section{A Simple Messaging System  for Interaction}
\label{sec:interact}
In this section, we describe a simple message-based
communication mechanism between a pair of agents. It consists of messages from a sending agent to a receiving agent.

\begin{mydefinition}[Messages]
A message  is a 3-tuple of the form $(a,m,b)$ where $a$
is the identifier of the sender; $b$ is the identifier of the
receiver; and $m$ is a finite length message-string. 
\end{mydefinition}

\noindent
We define the following categories of 
message-strings:
\begin{description}
\item[Termination.] The message-string  $m = ''\Box''$
	denotes that sender is terminating the communication
	with receiver.
\item[Question.] The message-string $m = ''?_n(s)''$ denotes
	that the sender is sending a question $s$ with identifier
	$n$ to receiver.
\item[Answer.] The message-string $m = ''!_n(s)''$ denotes that the
	sender is sending an answer $s$ with identifier
	$n$ to receiver.
\item[Statement.] The message-string $m = ''\top(s)''$ denotes that
	sender is sending a statement to receiver.
\end{description}

\noindent
In the rest of the paper, we omit the identifier n as long as the context is clear. In the above definitions, s represents a sequence of message strings. However,
in this paper, in $?(s)$ $|s| = 1$ (that is, only 1 question is allowed at a time).
There can be multiple answers or even no answers for a question. That is,
in $!(s)$, $|s| \geq 0$. Additionally, in $\top(s)$, we will require $|s| \geq 1$.
If ordering is unimportant, we will sometimes show $s$ as a set instead of a sequence.
In all cases, if $|s| = 1$, we will denote the message by the singleton element,
dispensing with the sequence (or set) notation.

\noindent
The interaction between a pair of agents consists of 1 or more turns.
\begin{mydefinition}[Interaction]
A (1-step) turn from agent $a$ to agent 
$b$ is the pair of messages $ (M_1,M_2)$, where
$M_1 = (a,m_1,b)$, $M_2 = (b,m_2,a)$, and $m_1 \neq \Box$.
A k-step turn is the sequence $\langle T_1,T_2,\cdots,T_k \rangle$,
where $T_i$ $(1 \leq i \leq k)$ is a 1-step turn from $a$ to $b$.
Similarly for 1-step and $k$-step turns from $b$ to $a$.
We will call a sequence of 1 or more turns between $a$ and
$b$ an interaction between $a$ and $b$.
\end{mydefinition}

\begin{myexample}[Interaction]\label{ex:hm1}
Let $h$ denote a human agent and $m$ denote a machine.
Below is an example of a possible 3-step turn interaction between $h$ and $m$: \\
{
$\langle$ 
$(h,\top(\mathrm{Child~x~has~a~height~is~4~ft.}),m)$,$(m,\top(\mathrm{ok}),h)$,\\
$(h,\top(\mathrm{The~height~of~child~y~is~the~square~root~of~the~height~of~child~x}),m)$,
$(m,\top((\mathrm{ok}),h)$,\\
$(h,?_1(\mathrm{What~is~the~height~of~y}),m)$, $(m,!_1(\mathrm{y~is~+2~or~-2}),h)$
$\rangle$
}
\end{myexample}

\begin{myexample}[Interaction (contd.)]\label{ex:hm2}
A 4-step turn interaction between $h$ and $m$ is:\\
$\langle (h,\top(\mathrm{Child~x~has~a~height} \mathrm{is~4~ft.}),m)$,
$(m,\top(\mathrm{ok}),h)$,\\
$(h,\top(\mathrm{The~height~of~child~y~is~the~square~root~of~the~height~of~child~x}),m)$,$(m,\top(\mathrm{ok}),h)$,\\
$(h,?_1(\mathrm{What~is~the~height~of~y}),m)$,
$(m,!_1(\mathrm{y~is~+2~or~-2}),h)$,\\
$(h,\top(\mathrm{Your~answer~is~not~completely~correct~since~height~has~to~be~positive}),m)$,
$(m,!_1(\mathrm{y~is~+2}),h)$
$\rangle$
\end{myexample}

We note that each turn consists of a sequence of 2 messages.
Thus, with every interaction consisting of $k$ turns
$\langle T_1,\cdots,T_k\rangle$, there exists a corresponding
sequence $\langle M_1, M_2, \cdots, M_{2k} \rangle$ of messages
and $\langle m_1, m_2, \cdots, m_{2k} \rangle$ message-strings.
By construction, the sequence
$\langle m_1, m_3, \ldots, m_{2k-1} \rangle$ will be from agent $a$
to agent $b$, and
$\langle m_2, m_4, \ldots, m_{2k} \rangle$ will be from agent $b$
to agent $a$.
We denote these as $\langle m_{ab} \rangle$ and
$\langle m_{ba} \rangle$ for short. 

\noindent
Interaction sequences allow us to define the \textit{context} for
an agent. We assume any agent has access to a (possibly empty)
set of prior statements, which we call \textit{background knowledge}.

\begin{mydefinition}[Interaction Context]
\label{def:context}
Without loss of generality, let $(T_1,T_2,\cdots,T_k)$ be a k-step interaction
from agent $a$ to agent $b$. We denote the interaction context
for $a$, or simply the context for $a$, on the the $i^\text{th}$
turn $T_i$ as $C_{a,i} = $
$(m_1,m_2,\ldots,m_{2i-2})$, and the interaction context for $b$, 
, or simply the context for $b$, on the $i^\text{th}$ turn
$C_{b,i}$ $=$
$(m_1,m_2,\ldots,m_{2i-1})$. 
\end{mydefinition}
\noindent
In this paper, we are interested in $k$-step interactions between
agents $a$ and $b$, in which the context $C_{a,k}$ contains
a question $?_\alpha(q)$ and the context $C_{b,k}$ terminates
with a corresponding answer $!_\alpha(a)$. We define a special agent $\Delta$ called the
{\em oracle\/}. The oracle's answers are taken to be always correct.

\begin{myremark}[Interaction with the Oracle]
We note the following special features of the oracle:
\begin{itemize}
    \item $\Delta$ knows everything upto the    
        present, including the
        content of conversations between any non-oracular agents;
    \item Only a 1-step interaction is allowed between a non-oracular
        agent $a$ and $\Delta$. The interaction consists of a turn $T$ where:
        \begin{itemize}
            \item[$\rightarrow$] $T = ((a,?q,\Delta),(\Delta,!(s),a))$; or 
            \item[$\rightarrow$] $T = ((a,?q,\Delta),(\Delta,\Box,a))$.
        \end{itemize}
    \item The answer(s) provided by $\Delta$ are always correct.
\end{itemize}
\end{myremark}

\begin{myexample}[Interaction with an Oracle]\label{ex:ho1}
We consider the example again, this time with the 
human agent interacting with an oracle $\Delta$.
A 1-step interaction between $h$ and $\Delta$ is:\\
$\langle$
$((h,?_1(\mathrm{What~is~the~height}$ $\mathrm{of~child~y}),\Delta)$, $(\Delta,!_1((\mathrm{y=+2},\Box)),h))$
$\rangle$
\end{myexample}

\noindent
The oracle allows us to categorise questions and answers.

\begin{mydefinition}[Incomplete Question]
\label{def:pii}
Without loss of generality, let $(T_1,T_2,\cdots,T_k)$ be a k-step interaction
from $a$ to $b$. Let $C_{b,i}$ denote the context for $b$ on the $i^\text{th}$ turn. Let $T_i = ((a,?(q),b),\cdot)$, where agent a sends question q to b.
Let $((b,?(q),\Delta),(\Delta,!(s),b))$
be an interaction between $b$ and $\Delta$. If $s=\Box$, we conclude that  $q$ is incomplete.
In such a case, we will also say it is incomplete for $b$ given
$C_{b,i}$.
\end{mydefinition}

\noindent
That is, a question is incomplete, if the oracle does not give an answer. This is because
if the oracle cannot answer, neither can $b$. Similarly:

\begin{mydefinition}[Ambiguous Question]
\label{def:paa}
Without loss of generality, let $(T_1,T_2,\cdots,T_k)$ be a k-step interaction
from $a$ to $b$. Let $C_{b,i}$ denote the context 
for $b$ on the $i^\text{th}$ turn. Let $T_i = ((a,?(q),b),\cdot)$, where agent a sends question q to b.
Let $((b,?(q),\Delta),(\Delta,!(s),b))$
be an interaction between $b$ and $\Delta$. If $|s| > 1$ then we will say $q$ is ambiguous.
In such a case, we will also say $q$ is ambiguous for $b$ given $C_{b,i}$.
\end{mydefinition}

\noindent
That is, a question is ambiguous if the oracle returns more than one answer.
We assume that questions are either incomplete or ambiguous, but not both at once.

We are still left with the  requirement of needing to consult the oracle
to decide whether a question is one or the other. This is impractical for
several reasons: (1) Such an agent may not be available; (2) Even if available,
it may be extremely expensive to consult an oracle; and (3) Even if
costs were manageable, the oracle is not an ideal agent for inclusion
in the design of conversational system. It only engages in single-turn
conversations, does not provide any explanations for answers, and makes
no attempt to provide answers by resolving any incompleteness or ambiguity.
To address these issues, we introduce the notion of
\texttt{context transducers}.

\begin{mydefinition}[Context-Transducer]
Let $ {\cal C}$ denote the set consisting of sequences of message-strings.
A context-transducer agent is a function $\Lambda: {\cal C} \rightarrow {\cal C}$.
\end{mydefinition}

Our interest in this paper is on transducers that
operate on contexts that terminate in a question.

\begin{mydefinition}[Context-Transducer for Questions]
\label{def:qtrans}
Let ${\cal C}$ denote the set consisting of sequences
of message-strings. 
A context-transducer for questions, or simply a question-transducer, is a function
that maps any
$(m_1,m_2,\ldots,m_j,?_\alpha(q)) \in {\cal C}$
to some $(m'_1,m'_2,\ldots,m'_k,?_\alpha(q')) \in {\cal C}$.
\end{mydefinition}

Note that as defined, a question transducer can not only reformulate
the question, but it can also alter and augment the context string it received.\footnote{Clearly, a similar definition can be provided for
answer-transducers. We will not require it in this paper.}
Our focus in this paper will be on \texttt{context-preserving} and
\texttt{context-augmenting} transducers for questions. In
the former $j = k$ and $m'_i = m_i$ in Def. \ref{def:qtrans} above.
In the latter $j < K$ and $m'_i = m_i$ for $i = 1 \ldots j$.
In both cases, $q$ and $q'$ may differ.

Next we describe the implementation of a context-transducer for
questions that uses an agent-based architecture to identify
incomplete or ambiguous questions, and provide a reformulated
question and context.

\section{Implementation of a Question-Transducer}

Our implementation of a question-transducer will involve two functions:
(1) \texttt{Classify}: ${\cal C} \mapsto {\cal C} \times {\cal Y}$,
where ${\cal Y}$ is a set of labels $\{incomplete,ambiguous,normal\}$; and
(2) \texttt{Resolve}: ${\cal C} \times {\cal Y} \mapsto {\cal C}$. Given an
initial context $c \in {\cal C}$, the question-transducer computes
\texttt{Resolve}$($\texttt{Classify}$(c))$.\footnote{If $c$ does not terminate in a question,
then \texttt{Classify}$(\cdot)$ simply returns $(normal,c)$. If \texttt{Resolve}$(\cdot)$
receives an input of $(normal,c)$, it simply returns
$c$} 
The functions themselves are routine (Procedures
\ref{alg:classify} and \ref{alg:fresolve}). 

\begin{algorithm}[!htb]
\small
\caption{The \texttt{Classify} Function.}
\label{alg:classify}
    \textbf{Input}:
        $c$ : a context;
        $\Lambda_l$: a question-classifier;
        $B$: domain-specific information\; \\
    \textbf{Output}:  
        $(c',y)$, where $c'$ is a context, and $y' \in \{incomplete,ambiguous,normal\}$
    \begin{algorithmic}[1]
    \If{$c = (m_1, m_2, \ldots, m_k, ?_\alpha(q))$}
        \State $(y, e) := \Lambda_l(c,B)$ \Comment{$e$ is an explanation}
    \Else
        \State $y := normal$
    \EndIf
    \State \textbf{return} $(c, y)$
\end{algorithmic}
\end{algorithm}

\begin{algorithm}[!htb]
\small
\caption{The \texttt{Resolve} Function}
\label{alg:fresolve}

\textbf{Input:} $lc$: a labeled context;
    $\Lambda_r$: a question-resolver;
    $B$:  domain-specific information\;\\
\textbf{Output:} $c'$: a context

\begin{algorithmic}[1]
    \State Let $lc = (m_1, m_2, \ldots, m_k, ?_\alpha(q), y)$
    \If{$y = \text{incomplete} \ \vee \ y = \text{ambiguous}$}
        \State $(?_\alpha(q'), e) \gets \Lambda_r(c,B)$ \Comment{$e$ is an explanation}
        \State $c' \gets (m_1, m_2, \ldots, m_k, e, !_\alpha(q'))$
    \Else
        \State $c' \gets c$
    \EndIf
    \State \Return $c'$
\end{algorithmic}
\end{algorithm}

In this paper, $\Lambda_c$ and $\Lambda_r$ are implemented using
LLM-based \textit{agents}. Procedure \ref{alg:llm_agent} shows a 
a simplified form of a goal-based agent (in the sense
used \cite{brewka1996artificial}). $\Lambda_c$ and $\Lambda_r$ use instances
of this generic agent to achieve the goals of question-classification
and question-resolution (the generic agent includes some additional parameters,
which will need to be provided by the \texttt{Classify} and \texttt{Resolve} functions).

\begin{algorithm}[htb]
\small
\caption{A Goal-Based LLM Agent}
\label{alg:llm_agent}

\textbf{Input:}
 $\lambda$: an LLM model;
 $c$: a context;
 $B$: prior information;
 $g$: a goal\\
\textbf{Output:}
    $(r, e)$: $r$ is a response and
        $e$ is an explanation of how $r$ was obtained

\begin{algorithmic}[1]
    \State $C := c$
    \State $Done := False$
    \State $(r,e) := (\emptyset,\emptyset)$
    \While{$\neg Done$}
        \State $P := AssemblePrompt(C,B)$
        \State  $(r,e) := Call\_LLM(\lambda,P)$
        \If{$GoalAchieved(g,r) \vee (r = \emptyset)$}
            \State $Done:= True$
        \Else 
            \State $C := Update(C,(r,e))$
        \EndIf
    \EndWhile
    \State \Return $(r, e)$
\end{algorithmic}
\end{algorithm}

The functionality of Procedure \ref{alg:llm_agent} is in fact readily
available in recent implementations of LLM-based \textit{agentic}
software systems. We show by way of example how $\Lambda_c$,
the question-classifier can be implemented using a widely
used LLM-agent software environment.

\begin{myexample}[Implementing $\Lambda_c$]
\label{ex:agent_classifier}
The question-classifier can be implemented using \texttt{LangChain} \cite{topsakal2023creating}
as follows. \texttt{Langchain} requires as input the following:
(a) An LLM model (correctly, an API to the model);
(b) A ``profile'', which is essentially a structured prompt that
contains the following:
    (i) The question;
    (ii)  The context preceding the question; 
    (iii) The goal of classifying the question as $incomplete$,
            $ambiguous$, or $normal$;
    (iv) The kind of agent-behaviour required (for example,
        a ``0-shot ReAct agent.''

Given these inputs, \texttt{LangChain} implements a (sophisticated)
version of the iterative procedure in Procedure \ref{alg:llm_agent}, and
returns as output a response and an associated explanation.
The output depends on the agent-behaviour specified. For example, when
configured as a 0-shot ReAct agent, the LLM specified by (a) is invoked with
the structured prompt, LLM analyses the question in relation to the context and classification goal. 
The LLM first performs some reasoning to achieve the goal and then acts (here,
the action is to classify the question). Iteration results from
\texttt{LangChain} interpreting the result of the action, and
adding``observations'' by reinvoking the LLM with an updated context.
This may result in the  LLM possibly repeating
its reasoning and acts again based on the updated context.
\end{myexample}

\noindent
The question-resolution $\lambda_r$ can similarly be implemented, simply
by changing the goal provided (\textit{``Resolve incomplete or ambiguous questions by rewriting them into clear and complete questions'}').

\begin{figure}[htb]
    \centering
    \begin{subfigure}[]{0.40\textwidth}
        \centering
        \includegraphics[height=3cm]{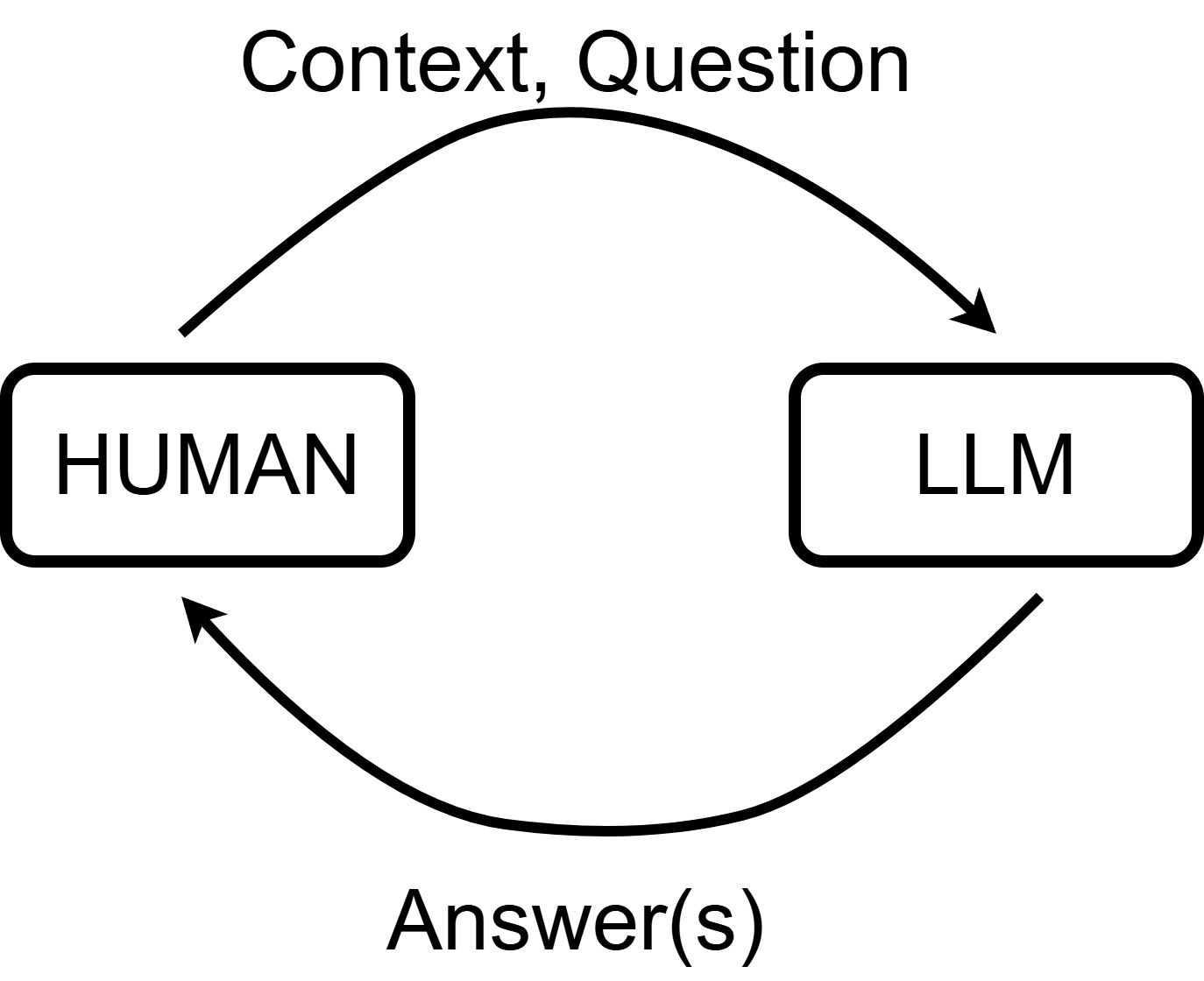}
        \caption{}
        \label{fig:qa_a}
    \end{subfigure}
    \hspace{0.05\textwidth}
    \begin{subfigure}[]{0.40\textwidth}
        \centering
        \includegraphics[height=3cm]{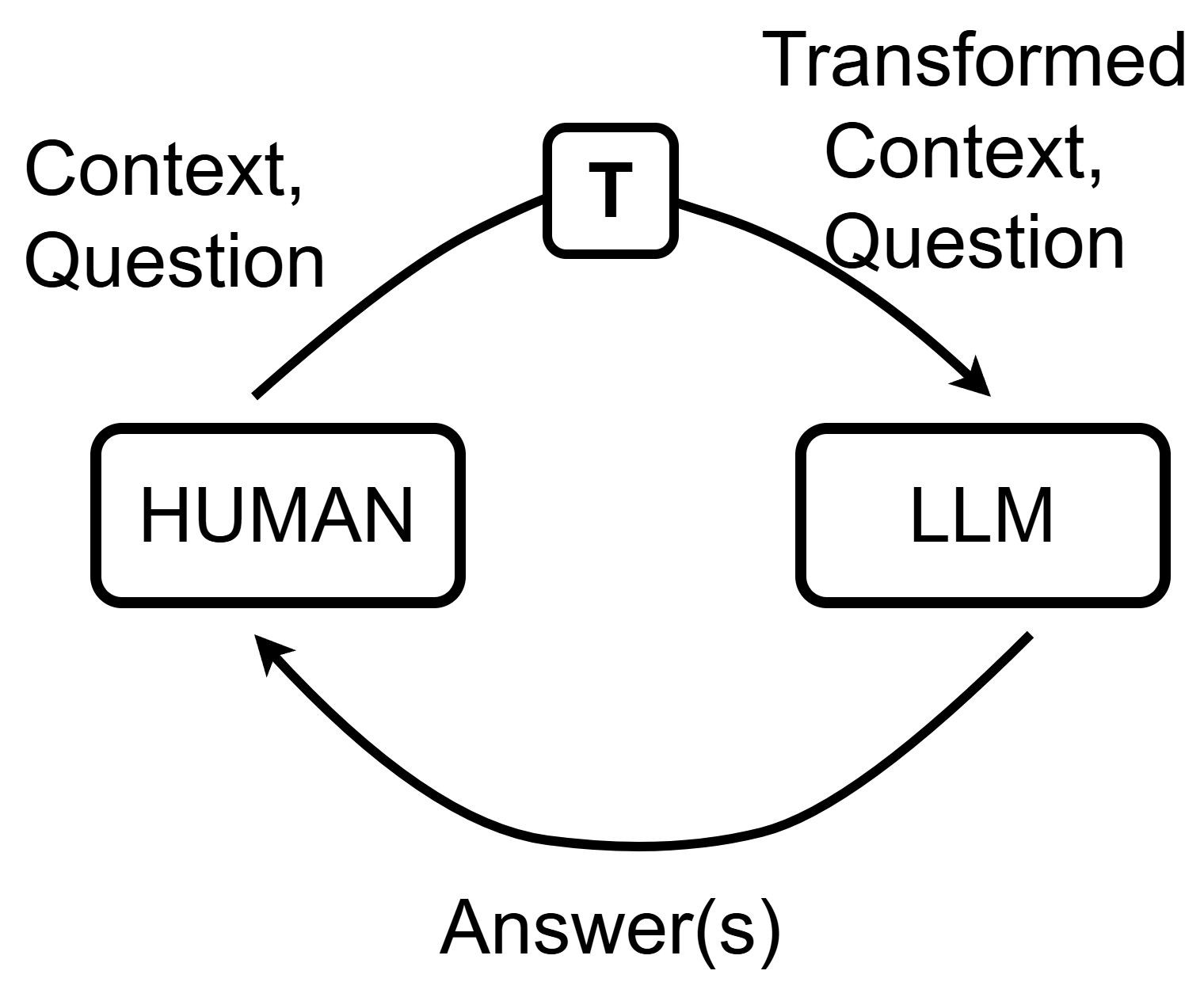}
        \caption{}
        \label{fig:qa_b}
    \end{subfigure}
    \caption{Human-LLM question-answering: (a) Without a question-transducer; and (b) With a question-transducer. For clarity, we have separated the question
    from the rest of the context.}
    \label{fig:qa}
\end{figure}

We now consider investigating empirically two variants of a basic interactive
question-answering system (Fig.~\ref{fig:qa}). We note that there
is no requirement for the LLM in Fig.\ref{fig:qa} to be the
same as the LLM used by the agent implementing the transducer.
To avoid confusion, we will call the LLM providing the answers
as the ``responder-LLM'' and the LLM used by the transducer
as the ``transducer-LLM''.

\section{Empirical Evaluation}
\label{sec:expt}

\subsection{Aims}
\label{secLaims}
We intend to test the following conjecture:
\begin{itemize}
    \item The inclusion of a question-transducer to detect and resolve
        incomplete or ambiguous questions improves the interactive
        question-answering performance of an LLM.
\end{itemize}

\noindent
We measure performance by the
accuracy of the answer returned by the LLM (using
human-labeled ground truth) after each turn of interaction.
Since the transducer necessarily introduces a delay, we will ideally
require accuracy to increase after each turn.\footnote{
Or--showing some measure of tolerance--at least after the first turn.
However, even this may not be sufficient for
improved performance, since the increase in accuracy may not be substantial
enough to tolerate the delay. We comment on these issues
in the section on Results.}
We test the hypothesis using datasets with varying degrees of incompleteness and ambiguity, and with transducers
constructed with 2 different LLMs (with varying amounts of
pre-training and parameters).

\subsection{Materials}
\label{sec:mat}
\subsubsection*{Datasets}
In our empirical study, we evaluate the QA systems on 6 datasets
of QA pairs (some datasets also contain some
additional background context for each question), with
varying amounts of incompleteness and ambiguity in the questions. 
We characterise the datasets into the following categories
(Appendix \ref{app:defs} describes how we arrive at this categorisation):

\noindent
\textbf{Category C1 (Low Incompleteness, Low Ambiguity).} SQuAD (Stanford 
    Question Answering Dataset):  a widely used 
    dataset for machine reading comprehension, consisting of over 100,000 
    questions based on Wikipedia \cite{rajpurkar-etal-2018-know}.
        
\noindent
\textbf{Category C2 (Low Incompleteness, Mid-to-High Ambiguity)} NQ-open (Natural 
    Questions open):  a large-scale benchmark, 
    featuring open domain user queries and answers annotated from Wikipedia 
    articles \cite{kwiatkowski-etal-2019-natural}; and
    AmbigNQ: dataset designed to handle ambiguous questions covering 14,042 NQ-open questions\cite{min-etal-2020-ambigqa,kwiatkowski-etal-2019-natural}. 

\noindent
\textbf{Category C3 (Mid-to-High Incompleteness, Low Ambiguity)}
    MedDialog: covering 0.26 million conversations between 
    patients and doctors curated to understand real-world medical queries 
    \cite{zeng-etal-2020-meddialog}.

\noindent
\textbf{Category C4 (Mid-to-High Incompleteness, Mid-to-High Ambiguity)}
    MultiWOZ (Multi-domain Wizard-of-Oz): a dataset covering 
    multiple domains such as hotels, restaurants, and taxis with 8438 dialogues \cite{budzianowski-etal-2018-multiwoz}; and
    ShARC (Shaping Answers with Rules through Conversation): a multi-turn 
    dataset that focuses on 32k task-oriented conversations with reasoning 
    covering multiple domains\cite{saeidi-etal-2018-interpretation}.

\noindent
As explained in Appendix \ref{app:defs}, the rules used to obtain
these categories can only be used to classify interaction sequences,
and therefore cannot be used by a question-transducer during the
course of an interaction.
Using the terminology introduced in
Sec.~\ref{sec:interact}, each dataset can be seen
as a set of interactions, with each interaction being a sequence 
1 or more turns. 
SQuAD, MultiWOZ and ShARC include some relevant context along
with the question in the first turn.
MultiWOZ and MedDialog are interactions between human and LLM agents.
In all other cases, interactions are between human agents.

\subsubsection*{Algorithms and Machines}
We use the following Large Language Models and software: (a) GPT3.5-Turbo and Llama-4-Scout: LLMs are used to retrieve the answer to the user questions. (b) LangChain: is used to invoke LLMs and initiate agents. All implementations are in Python 3.10, with API calls to the respective model engine. Experiments are conducted on a workstation based on Linux (Ubuntu 22.04) with 256GB of RAM, an Intel i9 processor, and an NVIDIA A5000 graphics processor with 24GB of memory. 

\subsection{Method}
\label{sec:method}

We first clarify some preliminary details:
(a) Recall data-instances are sets of interactions,
    with each interaction being a sequence of 1 or more turns.
    The first turn containing the question; and the last turn
    containing the final answer. Additionally, some datasets have
    prior context information relevant to the question. For
    our purpose, it is sufficient to represent each interaction $I$ 
    as $(c,r)$ pairs, where
    $c$ represents a context sequence $(m_1,m_2,\ldots,m_j)$ where
    $m_1,\ldots,m_{j-1}$ denotes any prior information available 
    and  $m_j = ?_\alpha(q)$ is a question. For datasets where no prior
    information is available, $j=1$. The response $r$ 
    will be a sequence $(!_\alpha(a))$, containing
    the final answer in $I$;
(b) We assume the functions $Q_\alpha(c)$ and $A_\alpha(c)$ return
    the question $?_\alpha(q)$ and answer $!_\alpha(c)$ in
    any sequence of messages if they exist, or $\Box$ otherwise;
(c) We assume a responder-LLM $\lambda$ that provides
    a response to a context $c$, and denote this by $\lambda(c))$.

With these preliminary clarifications aside, our method is
straightforward:
\begin{enumerate}
    \item For turns $k = 1,2,\ldots,K$ do:
        \begin{enumerate}
            \item For each dataset $d$:
            \begin{enumerate}
                \item Obtain
                        $C_{d,k} = \{(c,r): (c,r) \in d, q=Q_\alpha(c), a=A_\alpha(r), a_\lambda = A_\alpha(\lambda(c)), \mathtt{AGREES}(a,a_\lambda)\}$ be the
                        set of responses where the answer from the LLM
                        agrees with that in the dataset
                \item  $C_{T,d,k} = \{(c,r): (c,r) \in d, q = Q_\alpha(c), a=A_\alpha(r), a_{T,\lambda} = A_\alpha(\lambda(T(c))), \mathtt{AGREES}(a,a_{T,\lambda})\}$ be the
                        set of responses where the answer from the LLM
                        using the transduced context agrees with that in the dataset
                \item Calculate ${Acc}_{d,k} = |C_{d,k}|/|d|$ (the accuracy without the transducer); and ${Acc_{T,d,k} = |C_{T,d,l}|/|d|}$, the
                    accuracy with the transducer
            \end{enumerate}
        \end{enumerate}
\end{enumerate}

\noindent
The following additional details are relevant:
\begin{itemize}
    \item For the experiments reported here,  $K = 3$;
    \item The responder-LLM in all cases is $GPT-3.5 Turbo$;
        \item The question-transducer is implemented using
            \texttt{LangChain} agents for question-classification
            and resolution. In both cases, the agent is
            configured to be a 0-shot ReAct agent (see Example \ref{ex:agent_classifier}). It is desirable for the transducer-LLM
            to be as lightweight
            possible (to decrease delays). 
            We report on results with $GPT-3.5-Turbo$ and $Llama-4-Scout$ as
            the transducer-LLM;
        \item The $AGREES$ relation used to obtain the sets of
        correctly answered questions is a manual verification step,
        since we believe this to be most reliable to obtain the
        statistics we require. It can however entail
        manual verification of upto 1800 responses (there are 600
        $(c,r)$ entries overall; and we are examining responses upto
        $K=3$ turns.
\end{itemize}

\subsection{Results}
\label{sec:results}
The principal results from the experiment is
in Table \ref{tab:results}. A summary of the main findings from the table is:
(1) For 5 of the 6 datasets, introducing the transducer increases accuracy
    (even from the first turn);
(2) The gain in accuracy through the use of the transducer ranges from
    $3\%$ to $40\%$ by the last turn tabulated; and
(3) In 3 datasets the GPT-based
    transducer appears to perform better than the Llama-based transducer; and
    \textit{vice versa}.
Taken together, we believe these results provide strong support for
the experimental conjecture of the use of a transducer being beneficial.

\begin{table}[h!]
\centering
\small
\caption{Accuracy across datasets comparing QA with and without a transducer. Role of increasing context on the proportions of correct answers.}

\begin{tabular}{|l|c|c|c|c|c|}
\hline
\textbf{Dataset} & \textbf{Turns} & \multicolumn{3}{c|}{\textbf{Accuracy}}\\ 
& & \textbf{Without } & \multicolumn{2}{c|}{\textbf{With Transducer}}\\
& & \textbf{Transducer} & \\ 
& & & \textbf{GPT-based} & \textbf{LLaMA-based} \\
\hline
\multirow{3}{*}{SQuAD}
& 1 & 0.92 & 0.71 & 0.75 \\
& 2 & 0.95 & 0.94 & 0.80 \\
& 3 & 0.97 & 0.97 & 0.85 \\
\hline

\multirow{3}{*}{NQ-open}
& 1 & 0.81 & 0.83 & 0.80 \\
& 2 & 0.87 & 0.92 & 0.89 \\
& 3 & 0.89 & 0.94 & 0.93 \\

\hline
\multirow{3}{*}{AmbigNQ}
& 1 & 0.63 & 0.65 & 0.59 \\
& 2 & 0.69 & 0.86 & 0.74 \\
& 3 & 0.78 & 0.90 & 0.81 \\

\hline
\multirow{3}{*}{MedDialog}
& 1 & 0.00 & 0.08 & 0.15 \\
& 2 & 0.18 & 0.36 & 0.41 \\
& 3 & 0.26 & 0.50 & 0.66 \\

\hline
\multirow{3}{*}{MultiWOZ}
& 1 & 0.04 & 0.22 & 0.24 \\
& 2 & 0.25 & 0.49 & 0.52 \\
& 3 & 0.48 & 0.57 & 0.69 \\

\hline
\multirow{3}{*}{ShARC}
& 1 & 0.11 & 0.19 & 0.12 \\
& 2 & 0.60 & 0.56 & 0.60 \\
& 3 & 0.83 & 0.82 & 0.85 \\

\hline
\end{tabular}
\label{tab:results}
\end{table}

\noindent
We now examine these results in more detail:

\noindent
\textbf{When the Transducer Does Not Help.} It is evident that 
the transducer does not help with the SQuaAD dataset. In our categorisation,
this is a dataset with low incompleteness and low ambiguity. That is, questions
are largely factual questions. Additionally, each question has substantial
additional context. In
other words, almost all questions are ``normal''. In principle, such
questions should pass through the transducer unchanged, but clearly the
drop in accuracy suggests that they do not. The problem appears to
lie in the \texttt{Classify} function implemented within the transducer.
We examine is next.

\noindent
\textbf{The \texttt{Classify} and \texttt{Resolve} Functions.} The label distributions
across datasets obtained from the \texttt{Classify} function used by the transducer
are in the Table \ref{tab:stats_classify} of Appendix \ref{app:tstats}. The tabulation shows that:
(a) the two transducer-LLMs behave very differently;
(b) contrary to expectations, neither transducer-LLMs find many
    questions to be ambiguous (this is even in a dataset like
    AmbigNQ, which has been curated to include ambiguous questions; and
(c) The proportions of incomplete and ambiguous questions are quite different
    to the proportions from the retroactive classification obtained using
    the rules in Appendix ~\ref{app:defs}. We observe that without \texttt{Classify} the accuracy drops.
    These 3 observations make it unclear what form of incompleteness
    of ambiguity is actually being captured by the agent-based
    implementation. To assess the performance of the \texttt{Resolve} function, we
need to estimate how often a resolution improved accuracy. This information
is tabulated in Table \ref{tab:stats_resolve} of Appendix \ref{app:tstats}, and shows probability of
improving the accuracy of an answer is substantially greater if
\texttt{Resolve} function transforms the question. This suggests that
\texttt{Resolve} is being implemented substantially more along expected
lines than \texttt{Classify}.

\noindent 
\textbf{Cost of Using a Transducer.} Introduction of the transducer introduces 
costs in the form of increased calls to the transducer-LLM (and associated
time-delays). Table \ref{tab:stats_api} in Appendix \ref{app:tstats} contains estimates
of these for each dataset. In general, the Llama-based transducer appears
to make more calls to the transducer-LLM than GPT. This
suggests that using the Llama-based transducer may be costlier--by way
of increased latency. This appears to be more likely for datasets in C3 and C4. 

\section{Related Work}
\label{sec:relwork}


Preliminary research in QA systems has focused on simple interactions, while this worked adequately for single-turn queries, they lacked context management required for engaging conversations \cite{na-etal-2022-insurance}. Deep learning \cite{zhao2018multi} and transformers \cite{radford2018improving,kenton2019bert} could improve the precision and coherence of interactions. However, maintaining a long context, handling deficient queries, and aligning responses with user intent still require attention. These issues underscore the need for a detailed study of human-machine interactions. Recent advances in LLM are changing the story with approaches to handle human-machine interactions \cite{huo2023retrieving,tan2023can}. Researchers are exploring interaction at its crux, focusing on differences between single-turn and multi-turn interaction \cite{zaib2022conversational,burggraf2022preferences,li2023s2m}. A study by \cite{srinivasan2024implementation} introduces intelligible interactions for improving answer quality and proposes a structured protocol for human–LLM communication. While these studies offer valuable insights, research on understanding and defining interactions at the grassroots in the era of LLM remains underexplored.

\noindent
{\bf Incompleteness and Ambiguity.} LLMs are designed to process and generate human-like text, yet they can still misinterpret context and provide an inscrutable answer due to deficiencies in the human queries and contextual information. Providing clarifying feedback helps keep the dialog accurate and easy to understand. To identify when to ask for feedback, we need to detect when a question is incomplete or ambiguous, i.e, when it contains deficiencies.
To understand incompleteness in questions, \cite{addlesee2023understanding} parse incomplete questions and repair them using human recovery strategies. They defined queries with missing entities as incomplete questions. Similarly, a study by \cite{kumar2017incomplete} considers an incomplete question as one that lacks either of the three: topic presentation, adjective expression, or the inquiry. A series of techniques are employed to detect ambiguity in the questions \cite{gao-etal-2021-answering}. Graphical information is used to define the ambiguity of a query as concepts covered and the dissimilarities between the answer and the query \cite{banerjee2021detecting}. A few studies frame it as scope ambiguity when a sentence contains multiple operators and overlapping scope \cite{kamath-etal-2024-scope}. LLMs are not predominantly trained to detect and resolve deficiencies, but can be instructed by injecting patterns derived from human-annotated data \cite{papicchio2024evaluating}. \cite{mehrab2025detecting} use LLMs to detect ambiguous queries using taxonomies and rewrite the queries in an enterprise setting. \cite{cole-etal-2023-selectively} focus on estimating LLM (Palm) confidence when answering ambiguous questions and employ prompt-based disambiguation, which proves to be sufficiently ineffective. \cite{kim-etal-2024-aligning} propose leveraging intrinsic knowledge. It introduces a measure that suggests whether the model considers input as ambiguous based on a pre-defined threshold. Researchers also evaluate few-shot disambiguation using rephrasing and adding context, finding that adding context improves the performance \cite{keluskar2024llms}. These approaches, though cost-effective, do not guarantee the accuracy and robustness as compared to fine-tuning. Moreover, fine-tuning itself comes with substantial training and data curation effort. Our paper builds upon similar veins, using agents to bridge the gap, identify and reduce the concept of incompleteness and ambiguity in natural language interactions.  

\section{Conclusion}
\label{sec:concl}
Our interest is in question-answering (QA) systems in natural language. While the
use of natural language has long been a desirable requirement for interfaces to
computers, the possibility of achieving it has proved significantly harder until the
advent and development of large language models (LLMs). Even at this early stage of
LLM technology, we are able to focus on aspects of specific issues that arise with
the use of natural language, within the ambit of QA systems. In this paper, we 
have concentrated on two such aspects: incompleteness and ambiguity. Both feature
during the course of an interaction in natural language, and we have
proposed a simple transducer mechanism for detecting and mitigating
these deficiencies in questions posed to an LLM. The transducer
is implemented by LLM-based agents that perform the functions of detection
and resolution of questions with incompleteness or ambiguity. Our results
show that in almost all datasets we tested, the
inclusion of the transducer can substantially improve
accuracy of answers and shorten clarificatory interactions. The
transducer does not appear to help if the data contains sufficient
context information and the questions do not contain much by
way of incompleteness or ambiguity. 

\subsection*{Limitations}
While our experiments demonstrate the potential of agent-based augmentation to improve question-answering, but they also point to some aspects that
need improvement. First, our assessment
of the responses in all cases is by human verifier. Repetition by other diverse
human-verifiers
would both mitigate bias and reduce variability. Second, the agentic pipeline requires multiple API calls per query, which results in increased latency and computational cost. It is helpful to therefore only invoke the transducer
when it is necessary. As we saw in Sec.~\ref{sec:results}, the agent-based
implementation of the \texttt{Classify} function seems quite unpredictable, and
in some cases counter-intuitive. It would seem better to construct a standard
discriminative model for \texttt{Classify}. Finally, we have focused on the
construction of the transducer: it would be useful to see if the
benefits observed from the use of the transducer also carry over
to other responder-LLMs.

\bibliography{main}

\appendix
\section{Characterisation of Datasets}
\label{app:defs}

 We propose the following tests to
characterise datasets based on the possible presence of
incompleteness and ambiguity using question-answer
pairs. We use the terminology and notation
in Sec.~\ref{sec:interact}.

\begin{mydefinition}[Possibly-Incomplete Question]
\label{def:pi}
Let database $d$ contain an entry
$I = (T_1,T_2,\cdots,T_k)$, denoting a k-step interaction between $a$ and $b$.
Let  $T_i = ((a,?_\alpha(q),b),(b,?_\beta(s1),a))$; and
$T_{i+1} = (a,!_\beta(s2),b),(b,s3,a))$, where $s3$ can be any statement. Then
we will say $q$ is a possibly incomplete question.
\end{mydefinition} 

\begin{mydefinition}[Possibly-Ambiguous Question]
\label{def:pa}
Let database $d$ contain en entry 
 $I = (T_1,T_2,\cdots,T_k)$, denoting a k-step interaction between $a$ and $b$.
Let $T_i = ((a,?_\alpha(q),b),(b,!_\alpha(s1),a))$; and
$T_{i+1} = (a,\top(s2),b),(b,s3,a))$. Then
we will say $q$ is a possibly ambiguous question.
\end{mydefinition} 

\noindent
We note the following about these definitions:
\begin{itemize}
\item The definitions cannot be used prospectively during an interaction
    to classify questions since they require information about the
    response. Thus, these cannot be used to implement the
    \texttt{Classify} function of the question-transducer.
    \item We note that the definitions for detecting possible incompleteness and
    possible ambiguity require interactions of more than 1 step, i.e., multi-turn interactions. The definitions
    can be generalised to span more than 2 turns, but the existence of 
    at least 2 turns demonstrating the pattern shown is taken to be sufficient here;
\item It is possible that $b$ may send $a$ questions to clarify further. This may
    result in that subsequent question itself being incomplete or ambiguous. In this
    paper, we will be concerned only with the initial question from $a$ to $b$.
\end{itemize}

\noindent
We can use the definitions to arrive at an estimate of
the proportion of interactions in each dataset that contain
possibly-incomplete and possibly-ambiguous questions, tabulated in Table \ref{tab:dataset_analysis}.
\begin{table}[h!]
\centering
\caption{Proportions of incomplete and ambiguous questions present in the datasets.}
\begin{tabular}{|l|c|c|c|c|}
\hline    
                &                   & \multicolumn{2}{|c|}{\bf{Questions}} \\
\textbf{Category} & \textbf{Dataset} & \textbf{Poss. Incomp.} & \textbf{Poss. Ambig.}\\
\hline
C1  & SQuAD & 0.00 & 0.08 \\
\hline   
C2 & NQ-open & 0.02 & 0.17 \\
& AmbigNQ & 0.01 & 0.36 \\
\hline
C3 & MedDialog  & 0.92 & 0.08  \\
\hline
C4 & MultiWOZ  & 0.21 & 0.75 \\
& ShARC & 0.28 & 0.61  \\ 
 \hline
\end{tabular}
\label{tab:dataset_analysis}
\end{table}

\section{Transducer Details}
\label{app:tstats}
\subsection{Statistics}
\begin{table}[H]
\centering
\caption{Classification of questions into incomplete, ambiguous, and normal by \texttt{Classify} at each turn.}
\begin{tabular}{|l|c|p{0.05\columnwidth}p{0.05\columnwidth}p{0.05\columnwidth}|p{0.05\columnwidth}p{0.05\columnwidth}p{0.05\columnwidth}|}
\hline
\textbf{Dataset} & \textbf{Turns} & \multicolumn{3}{|c|}{\textbf{GPT-based}} & \multicolumn{3}{|c|}{\textbf{LLaMA-based}} \\
 &  & \textbf{Incomp.} & \textbf{Ambig.} & \textbf{Nor.} & \textbf{Incomp.} & \textbf{Ambig.} & \textbf{Nor.} \\
\hline
\multirow{3}{*}{SQuAD}      & 1 & 0.73 & 0.03 & 0.24 & 0.29 & 0.05 & 0.66\\
                            & 2 & 0.20 & 0.01 & 0.35 & 0.14 & 0.03 & 0.74\\
                            & 3 & 0.05 & 0.00 & 0.36 & 0.05 & 0.01 & 0.88\\
\hline
\multirow{3}{*}{NQ-open}     & 1 & 0.89 & 0.02 & 0.09 & 0.16 & 0.03 & 0.81 \\
                     & 2 & 0.06 & 0.00 & 0.17 & 0.06 & 0.00 & 0.95 \\
                     & 3 & 0.02 & 0.00 & 0.19 & 0.02 & 0.00 & 0.96\\
                     \hline
\multirow{3}{*}{AmbigNQ}     & 1 & 0.48 & 0.04 & 0.48 & 0.10 & 0.07 & 0.83 \\
                     & 2 & 0.16 & 0.04 & 0.63 & 0.10 & 0.03 & 0.87\\
                     & 3 & 0.06 & 0.00 & 0.71 & 0.04 & 0.00 & 0.91\\
                     \hline
\multirow{3}{*}{MedDialog}   & 1 & 0.99 & 0.00 & 0.01 & 0.95 & 0.00 & 0.05 \\
                     & 2 & 0.90 & 0.01 & 0.01 & 0.77 & 0.04 & 0.09 \\
                     & 3 & 0.55 & 0.00 & 0.02 & 0.53 & 0.06 & 0.10 \\
                     \hline
\multirow{3}{*}{MultiWOZ}    & 1 & 0.83 & 0.02 & 0.15 & 0.52 & 0.01 & 0.47 \\
                     & 2 & 0.45 & 0.01 & 0.43 & 0.33 & 0.02 & 0.75\\
                     & 3 & 0.29 & 0.00 & 0.60 & 0.21 & 0.02 & 0.83\\
                     \hline
\multirow{3}{*}{ShARC}       & 1 & 0.60 & 0.39 & 0.01 & 0.82 & 0.05 & 0.13\\
                     & 2 & 0.62 & 0.03 & 0.17 & 0.36 & 0.01 & 0.49 \\
                     & 3 & 0.41 & 0.01 & 0.39 & 0.11 & 0.01 & 0.56 \\
\bottomrule
\end{tabular}
\label{tab:stats_classify}
\end{table}
\noindent

\begin{table}[H]
\centering
\small
\caption{Proportions of incomplete and ambiguous questions that were correctly resolved using \texttt{Resolve} with accurate answer at each turn.}
\begin{tabular}{|l|c|c|c|}
\hline
\textbf{Dataset} & \textbf{Turns} & \multicolumn{2}{|c|}{\textbf{Resolved and Accurate} }\\
 & &\textbf{GPT-based}&\textbf{LLaMA-based}  \\
\hline
\multirow{3}{*}{SQuAD}       & 1 & 0.50 & 0.20 \\
                     & 2 & 0.62 & 0.23 \\
                     & 3 & 0.64 & 0.24 \\
                     \hline
\multirow{3}{*}{NQ-open}     & 1 & 0.73 & 0.14 \\
                     & 2 & 0.75 & 0.16 \\
                     & 3 & 0.80 & 0.16 \\
                     \hline
\multirow{3}{*}{AmbigNQ}     & 1 & 0.33 & 0.07 \\
                     & 2 & 0.44 & 0.15 \\
                     & 3 & 0.45 & 0.16 \\
                     \hline
\multirow{3}{*}{MedDialog}   & 1 & 0.08 & 0.13 \\
                     & 2 & 0.44 & 0.37 \\
                     & 3 & 0.57 & 0.62 \\
                     \hline
\multirow{3}{*}{MultiWOZ}    & 1 & 0.19 & 0.08 \\
                     & 2 & 0.36 & 0.14 \\
                     & 3 & 0.36 & 0.21 \\
                     \hline
\multirow{3}{*}{ShARC}       & 1 & 0.19 & 0.10 \\
                     & 2 & 0.43 & 0.27 \\
                     & 3 & 0.67 & 0.36 \\
\bottomrule
\end{tabular}
\label{tab:stats_resolve}
\end{table}

\begin{table}[H]
\centering
\small
\caption{Average number of LLM invocations for each interaction in the dataset.}
\begin{tabular}{|l|c|c|}
\hline
\textbf{Dataset} & \multicolumn{2}{|c|}{\textbf{LLM Calls} }\\
 & \textbf{GPT-based}&\textbf{LLaMA-based}  \\
\hline
\textbf{SQuAD}       & 4 & 4 \\
\hline
\textbf{NQ-open}     & 3 & 5 \\
\hline
\textbf{AmbigNQ}     & 4 & 6 \\
\hline
\textbf{MedDialog}   & 4 & 5 \\
\hline
\textbf{MultiWOZ}    & 3  & 7  \\
\hline
\textbf{ShARC}       & 3 & 6 \\
\hline
\end{tabular}
\label{tab:stats_api}
\end{table}


\subsection{The Transducer-LLM}
We devise LLM-agents to \texttt{Classify}, \texttt{Resolve} user questions, and to generate response(s). The instructions for each are given below:

\begin{tcolorbox}[
  colback=gray!5,
  colframe=gray!30,
  title=Instructions to Classifier,
  fonttitle=\bfseries,
  coltitle=black,
  boxrule=0.5pt,
  arc=2pt,
  left=5pt,
  right=5pt,
  top=5pt,
  bottom=5pt
]
You are a classifier responsible for analyzing a question strictly based on the provided context and human clarification. Your job is to classify the question into one of the following categories:
``Normal'', ``Incomplete'', and ``Ambiguous''.
Make your classification solely based on the question, the human-provided clarification, and the given context. Include a brief explanation for your classification.
Ensure the response is formatted exactly as: \\
\texttt{`Classification: <class label> \\ Explanation: <explanation>'}
\end{tcolorbox}

\begin{tcolorbox}[
  colback=gray!5,
  colframe=gray!30,
  title=Instructions to Resolver,
  fonttitle=\bfseries,
  coltitle=black,
  boxrule=0.5pt,
  arc=2pt,
  left=5pt,
  right=5pt,
  top=5pt,
  bottom=5pt
]
You are a helpful assistant. Your task is to resolve incomplete or ambiguous questions by rewriting them into clear and complete questions based on the provided context and explanation. 
If the question cannot be resolved, return a clarifying question instead.
\texttt{`Resolved: <resolved question>'}
\end{tcolorbox}

\begin{tcolorbox}[
  colback=gray!5,
  colframe=gray!30,
  title=Instruction to Generate Response,
  fonttitle=\bfseries,
  coltitle=black,
  boxrule=0.5pt,
  arc=2pt,
  left=5pt,
  right=5pt,
  top=5pt,
  bottom=5pt
]
Answer a complete or resolved question using the context and human clarification given. If the information is insufficient, respond with a request for more details.
Format your response as:\texttt{`Answer: <answer>'}
\end{tcolorbox}
\noindent
we present an instance along with the response generated by the LLM without the transducer (\textit{LLM Response}), followed by how the question is classified and resolved by the LLM-agent using the question-transducer (\textit{LLM-agent Response}):
\begin{table}[h!]
\sethlcolor{green}
\centering
\small
\setlength{\tabcolsep}{8pt}
\caption{Snippet from AmbigNQ dataset to handle complete question using LLM-agent}
\begin{tabular}{|p{0.15\columnwidth}|p{0.70\columnwidth}|}
\hline
\multicolumn{2}{|c|}{\textbf{Data Sample}}\\ \hline
\textbf{Question} & Who scored the music for How to Train Your Dragon? \\  \hline
\textbf{Ground Truth} & John Powell \\ \hline
\multicolumn{2}{|c|}{\textbf{LLM-agent Response}}\\ 
\hline
\textbf{Classify} & The question seems straightforward, but I should verify if it's complete and clear. I'll start by classifying the question. Action: Classify Question. Action Input: Who scored the music for How to Train Your Dragon?, ('\hl{complete}', 'The question is complete because it clearly asks for a specific piece of information (the person who scored the music) and provides enough context (the movie title "How to Train Your Dragon") for the question to be understood and answered.') \\ \hline
\textbf{Answer} & Thought: Since the question is classified as complete, I can proceed to answer it directly. Action: Answer Question. Action Input: Who scored the music for How to Train Your Dragon?', 'The music for the movie "How to Train Your Dragon" was scored by John Powell. He is a British composer of film scores, and his work on the film earned him critical acclaim. Powell's score perfectly complemented the movie's themes of adventure, friendship, and self-discovery. He went on to score the sequels as well, including "How to Train Your Dragon 2" and "How to Train Your Dragon: The Hidden World".')\\ \hline
\textbf{Final Answer} & John Powell scored the music for How to Train Your Dragon. \\
\hline
\end{tabular}
\label{tab:complete}
\end{table}
\end{document}